\documentclass[letterpaper, 10 pt, conference]{ieeeconf}  % Comment this line out if you need a4paper

\IEEEoverridecommandlockouts                              % This command is only needed if 
\usepackage{kotex}
\usepackage{textcomp}
\usepackage{multirow}
\usepackage{graphicx}
\usepackage{lipsum}
\usepackage{longtable}
\usepackage{tabularx}
\usepackage{dsfont}
\usepackage{booktabs}
\usepackage{dblfloatfix}
\usepackage{xcolor}
\usepackage{amsmath}
\usepackage{hyperref}
\usepackage{bbm}
\usepackage{makecell}
\usepackage{threeparttable}
\usepackage{lipsum}
\usepackage{colortbl}
\definecolor{lightyellow}{rgb}{1,1,0.8}
\definecolor{lightred}{rgb}{1,0.8,0.8}
\definecolor{lightorange}{rgb}{1,0.9,0.7}

\title{\LARGE \bf
GeomGS: LiDAR-Guided Geometry-Aware Gaussian Splatting \\ for Robot Localization
}

\author{Jaewon Lee$^{1}$, Mangyu Kong$^{1}$, Minseong Park$^{1}$, and Euntai Kim$^{1,*}$% <-this % stops a space
\thanks{$^{*}$is that the corresponding author}%
\thanks{$^{1}$All the authors with the School of Electrical and Electronic Engineering,  Yonsei University, Seoul 03722, South Korea, {\tt\small \{leejaewon, mangyu0929,msp922,etkim\}@yonsei.ac.kr}}%
}

\begin{document}

\maketitle
\thispagestyle{empty}
\pagestyle{empty}

%%%%%%%%%%%%%%%%%%%%%%%%%%%%%%%%%%%%%%%%%%%%%%%%%%%%%%%%%%%%%%%%%%%%%%%%%%%%%%%%
\begin{abstract}

Mapping and localization are crucial problems in robotics and autonomous driving. Recent advances in 3D Gaussian Splatting (3DGS) have enabled precise 3D mapping and scene understanding by rendering photo-realistic images. However, existing 3DGS methods often struggle to accurately reconstruct a 3D map that reflects the actual scale and geometry of the real world, which degrades localization performance. To address these limitations, we propose a novel 3DGS method called Geometry-Aware Gaussian Splatting (GeomGS). This method fully integrates LiDAR data into 3D Gaussian primitives via a probabilistic approach, as opposed to approaches that only use LiDAR as initial points or introduce simple constraints for Gaussian points. To this end, we introduce a Geometric Confidence Score (GCS), which identifies the structural reliability of each Gaussian point. The GCS is optimized simultaneously with Gaussians under probabilistic distance constraints to construct a precise structure. Furthermore, we propose a novel localization method that fully utilizes both the geometric and photometric properties of GeomGS. Our GeomGS demonstrates state-of-the-art geometric and localization performance across several benchmarks, while also improving photometric performance.

\end{abstract}

%%%%%%%%%%%%%%%%%%%%%%%%%%%%%%%%%%%%%%%%%%%%%%%%%%%%%%%%%%%%%%%%%%%%%%%%%%%%%%%%
\section{INTRODUCTION}

3D Gaussian Splatting (3DGS) has attracted significant interest from various fields, ranging from computer vision and robotics to AR/VR, and it is considered promising direction mapping and localization. Since the inception of 3DGS, numerous studies have been conducted to improve the performance of 3DGS, with most relying on Structure-from-Motion (SfM) \cite{schoenberger2016sfm} to reconstruct the 3D structure \cite{scaffoldgs, li2024dngaussian}. Recently, some works have leveraged range data from LiDAR \cite{yan2024street},~\cite{hwang2024vegs} or estimated depth \cite{Fu_2024_colfree} rather than solely relying on Structure-from-Motion (SfM) \cite{schoenberger2016sfm} to refine the scale and structure of 3DGS. In particular, \cite{zhou2024drivinggaussian} proposed to impose a simple distance constraint between initial LiDAR points and Gaussian points, aiming to maintain the structural consistency of 3DGS.

\begin{figure}[tb]
  \centering
  \includegraphics[width=\columnwidth]{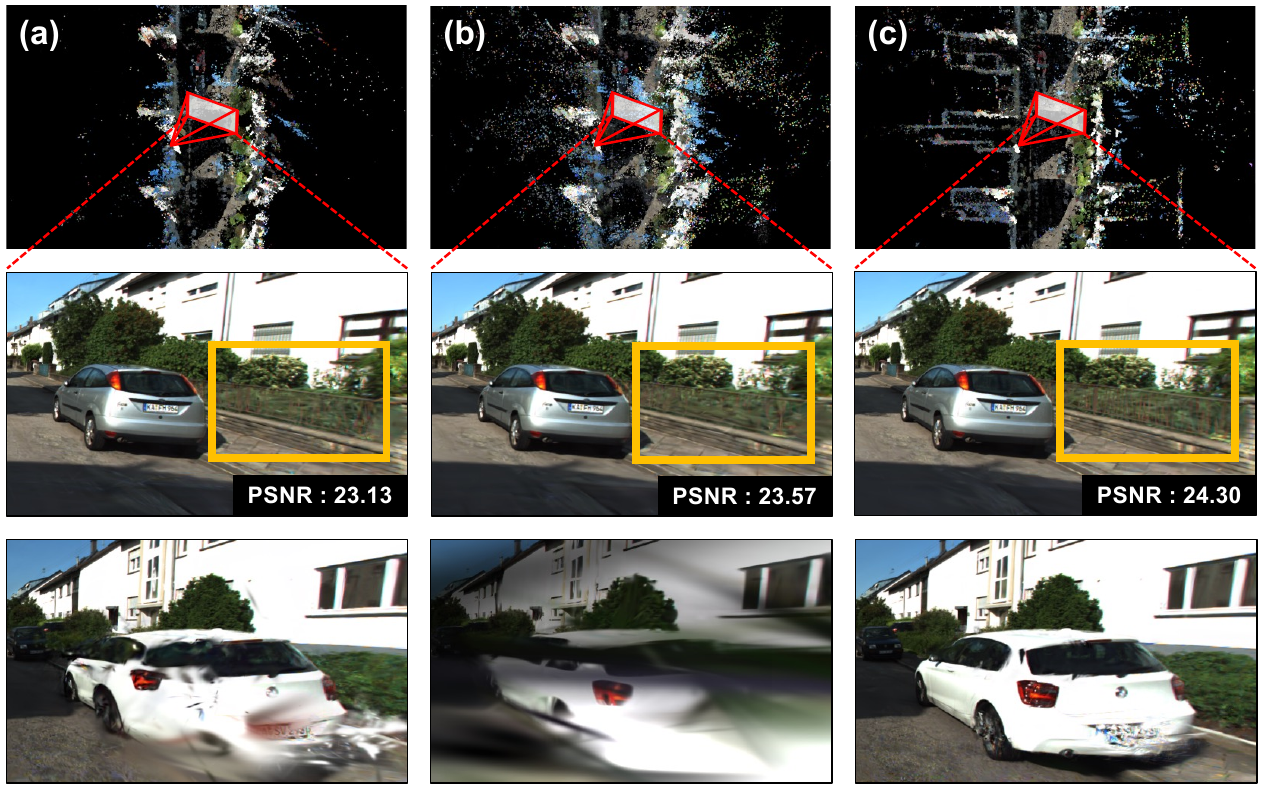}
  \caption{The qualitative results of GeomGS on the KITTI-360 dataset are as follows: (a) 3DGS created with SfM points, (b) 3DGS created with LiDAR points, and (c) \textbf{GeomGS}. The proposed method allows for observing finer details and can address cases where the structure is largely disrupted.}
  \label{figure_one}
\end{figure}

However, we believe there is still room for improvement in combining LiDAR with 3DGS to further enhance its quality. Specifically, the simple distance constraint developed in \cite{zhou2024drivinggaussian} focuses primarily on improving rendering quality, while relatively neglecting the geometric accuracy of points, potentially distorting the scale and structure of 3DGS depending on the environment. For instance, if this distance constraint is applied to the sky, enforcing the distances between LiDAR and 3DGS points to match could severely distort the geometric structure of the 3DGS, resulting in inaccurate 3D reconstruction.

To address the limitations of existing methods, we propose a novel Gaussian Splatting approach called Geometry-Aware Gaussian Splatting (GeomGS). GeomGS generates renderable maps that more accurately reflect real-world scales and structures. To achieve this, we introduce the Geometric Confidence Score (GCS), which evaluates the structural reliability of each Gaussian point. We incorporate probabilistic distance constraints~\cite{tian2020robust6d} based on the GCS, enabling the generation of a more accurate structure by focusing on higher-confidence Gaussian points while minimizing the influence of points that primarily affect image quality, thus preserving overall rendering quality.

Using these accurate renderable maps and the confidence of each Gaussian, we propose a new localization method that fully utilizes the rendering properties of 3DGS and the precise structure of GeomGS. We use a weighted Iterative Closest Point (ICP) \cite{original_icp} algorithm to align the query LiDAR scan within GeomGS by leveraging GCS values. Then, we optimize the pose by comparing the rendered image at the current pose with the ground truth image. We iteratively update these two methods. Through this integration, we design a robust and accurate localization technique.

We demonstrate the effectiveness of our method on various autonomous driving datasets, producing superior image quality and structurally accurate maps of the environment. Also, our approach shows significant improvements in localization accuracy compared to existing methods. Fig.~\ref{figure_one} presents qualitative results based on the initial points and shows that our method captures finer details and prevents scene degradation more effectively.

In conclusion, our proposed method makes the following contributions: \begin{itemize}

\item We introduce GeomGS, which uses a novel Geometric Confidence Score (GCS) and imposes probabilistic distance constraints between Gaussian and LiDAR points to generate geometrically accurate scenes. This enables the reconstruction of a map suitable for localization.

\item We propose a novel localization method integrating LiDAR-based localization with image-based pose optimization on a geometrically precise renderable map.

\item We comprehensively evaluate image quality, geometric accuracy, and localization performance, showing that our method achieves superior results. It outperforms existing techniques across various autonomous driving datasets.

\end{itemize}

\section{Related Work}
\subsection{Neural Scene Representation}
Recent advancements in novel-view synthesis and high-fidelity rendering have emerged from various approaches. Starting with NeRF \cite{mildenhall2020nerf}, improvements have been made in implicit representation through several notable works~\cite{barron2022mipnerf360},~\cite{barron2023zip},~\cite{tancik2022block}. An alternative and more advanced approach, 3D Gaussian Splatting (3DGS) \cite{kerbl3Dgaussians}, allows for real-time, point-based rendering, achieving superior image quality. The optimization of Gaussians for rasterization has led to advanced developments in novel-view synthesis. Following these advancements, recent methods~\cite{li2024dngaussian},~\cite{turkulainen2024dnsplatter},~\cite{cheng2024gaussianpro} leverage additional information such as image depth or image normals to achieve even higher-quality scene representations. Additionally, emerging works have explored the use of 2D~Gaussians instead of 3D~Gaussians to improve geometric accuracy in scene construction~\cite{Huang2DGS2024},~\cite{Dai2024GaussianSurfels}. Also, several works~\cite{martin2021nerfwild},~\cite{kulhanek2024wildgaussians},~\cite{lin2024vastgaussian} introduce various appearance models, to construct scenes that are robust to changes in lighting and environmental conditions. Commonly, these approaches use Structure-from-Motion (SfM) \cite{schoenberger2016sfm} to obtain initial points and camera poses. In our method, we improve structural accuracy and image quality by initializing points with LiDAR data and applying constraints and auxiliary approaches to enhance the results.

\subsection{Scene Reconstruction with Priors}
Recently, various NeRF and 3DGS works have introduced different types of priors or directly used point data to enhance scene reconstruction performance. In NeRF-based studies, notable works include S-NeRF \cite{ziyang2023snerf}, Point-NeRF \cite{xu2022point}, and Points2NeRF \cite{zimny2024points2nerf}. These studies focus on utilizing LiDAR point clouds or projecting LiDAR points onto images to build more accurate scenes. In 3DGS, which can directly manipulate point clouds, several works have been developed without relying on SfM. For instance, some approaches generate initial points using image information~\cite{turkulainen2024dnsplatter}, while others use NeRF results as priors \cite{niemeyer2024radsplat}. Further studies leverage LiDAR point clouds as initial points or apply simple distance constraints to reconstruct scenes~\cite{yan2024street},~\cite{hwang2024vegs},~\cite{zhou2024drivinggaussian}. Additionally, some methods utilize 3D Diffusion Models to generate initial point clouds \cite{yi2023gaussiandreamer}. Our work uses LiDAR point clouds as the initial points, applying probabilistic distance constraints and evaluating the geometric accuracy. We also introduce a novel method to make our map suitable for localization, addressing a new challenge in this field.

\subsection{Localization in Radiance Field}
Several works have explored localization and pose estimation using images within radiance fields, such as NeRF and 3DGS. In NeRF, numerous studies have demonstrated effective pose estimation by leveraging images~\cite{yen2020inerf}, particles~\cite{maggio2023loc},~\cite{kong2024fast}, and different optimization techniques~\cite{lin2023parallel},~\cite{bian2022nopenerf}. Similarly to our work, LocNDF \cite{wiesmann2023locndf} defines a Distance Field for efficient localization with LiDAR data. Additionally, a learning-based method \cite{chen2023dreg} is proposed for accurately registering NeRF blocks using surface fields. Recent works~\cite{sun2023icomma},~\cite{Bortolon20246dgs} with 3DGS, also utilize image-based pose estimation techniques. Building on these works, we propose a novel approach that fully utilizes rendering properties in radiance fields along with our point-based accurate map representation. This method demonstrates superior localization performance compared to existing approaches.

\section{METHODOLOGY}

   \begin{figure*}[tb]
      \centering
      \includegraphics[width=\textwidth]{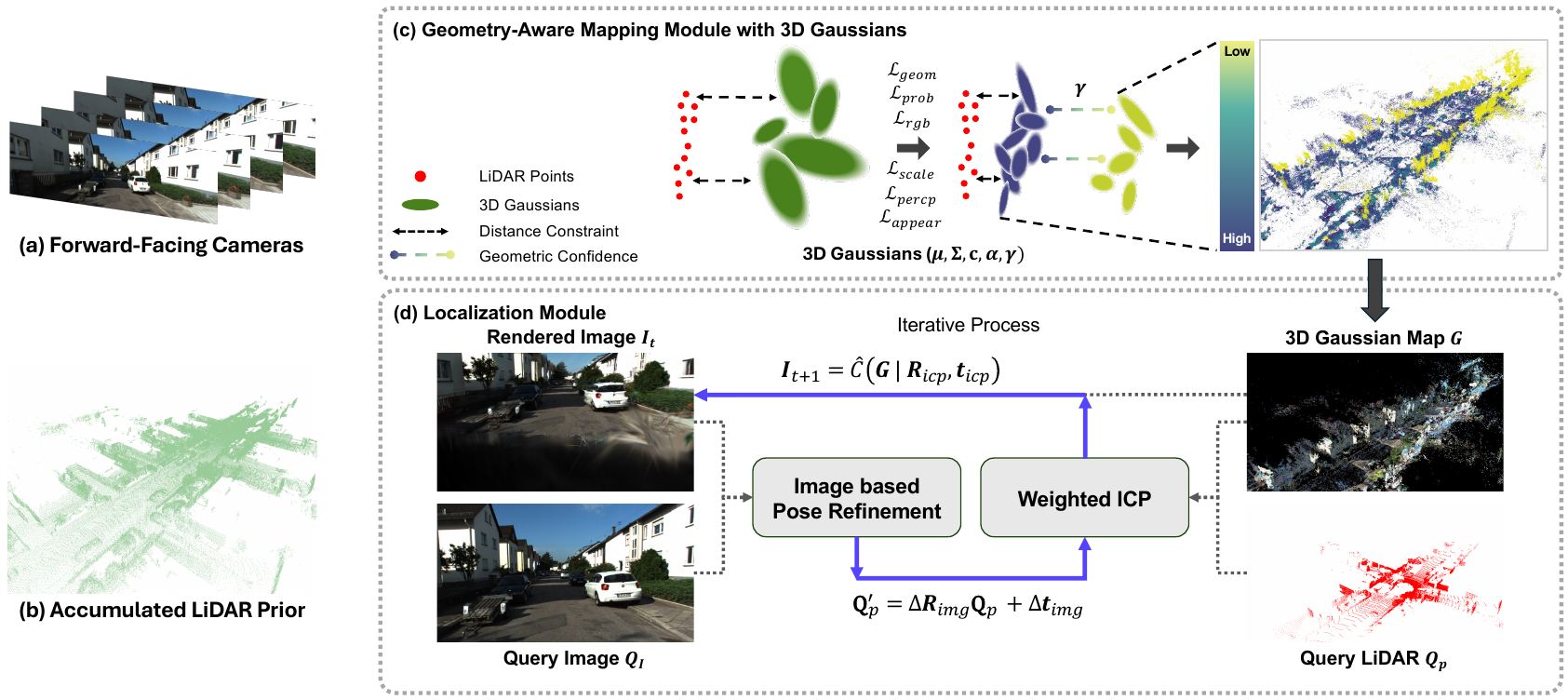}
      \caption{\textbf{Overall system of GeomGS.} (a), (b) We start with forward-facing images and poses from a dataset. The accumulated LiDAR points, based on the pose, are used as the initial points. (c) We perform geometrically accurate mapping. The parameters of the Gaussian are defined by mean, quaternion, color, and opacity. Additionally, the Geometrically Consistent Score (GCS) is used to identify points that are more geometrically reliable while remaining close to the given LiDAR points. (d) Our localization module fully utilizes LiDAR-based localization and renderable properties of Gaussians to perform iterative localization processes.}
      \label{figure_system}
   \end{figure*}
   
In this section, we introduce Geometry-Aware Gaussian Splatting (GeomGS). GeomGS integrates LiDAR data into conventional 3D Gaussian Splatting (3DGS) to significantly improve geometric accuracy and localization performance. Unlike existing methods, it efficiently utilizes LiDAR data to create an accurate map suitable for localization based on 3DGS. These improvements are particularly effective in applications such as autonomous driving and robotics. We start with a brief overview of 3DGS. We then introduce our Geometry-Aware Mapping method, which proposes a probabilistic distance loss based on the Geometric Confidence Score (GCS) to enhance structural and localization accuracy. Finally, we present our novel localization method, which leverages the precise geometry and photo-realistic rendering of 3DGS to improve localization in complex environments. Fig. \ref{figure_system} illustrates the overall system architecture, showing how LiDAR data is integrated throughout the process to enhance both mapping and localization.

\subsection{3D Gaussian Splatting with Real-Time Rendering}

In 3D Gaussian Splatting (3DGS) \cite{kerbl3Dgaussians}, each point in the scene is represented by 3D Gaussians, defined by its mean $\mu$, covariance matrix $\Sigma$, color $c$, and opacity $\alpha$. These Gaussians are flexible 3D primitives that are rendered efficiently by being rasterized into 2D. Each Gaussian primitive is represented as shown in Eq. \ref{basic_3dgs}:
\begin{equation}
\label{basic_3dgs}
G(x) = e^{-\frac{1}{2} (\mathbf{x} - \boldsymbol{\mu})^\top \mathbf{\Sigma}^{-1} (\mathbf{x} - \boldsymbol{\mu})}
\end{equation}
where $\mathbf{x}$ represents the 3D coordinates, and $\mathbf{\Sigma}$ is the covariance matrix that defines the shape and orientation of the Gaussian. The covariance matrix $\mathbf{\Sigma}$ can be computed using a rotation matrix $\mathbf{R}$ and a scaling matrix $\mathbf{S}$, as shown in Eq.~\ref{covmatrix}:
\begin{equation}
\label{covmatrix}
\mathbf{\Sigma} = \mathbf{R} \mathbf{S} \mathbf{S}^\top \mathbf{R}^\top
\end{equation}
Where $\mathbf{S}$ represents the scale of the Gaussian along each axis, and $\mathbf{R}$ defines its orientation in 3D space. Each Gaussian has a color $c$ and an opacity $\alpha$, which determine the appearance and transparency of the Gaussian during rendering. The final rendered image $\mathcal{I}$ can be represented by the following rendering function, as shown in Eq. \ref{render_func}:
\begin{equation}
\label{render_func}
\mathcal{I} = \hat{C}(\mathbf{G} \mid \mathbf{R}_c, \mathbf{t}_c) = \sum_{i=1}^{N} T_i \alpha_i c_i
\end{equation}
\begin{equation}
\alpha_i = (1-\text{exp}(-\sigma_i\delta_i)), \quad T_i = \prod_{j=1}^{i-1} (1 - \alpha_i)
\end{equation}
Here, $\hat{C}(\mathbf{G}\mid\mathbf{R}_c, \mathbf{t}_c)$ represents the rendered image~generated from a set of 3D Gaussians $\mathbf{G}$ under the camera pose defined by the rotation $\mathbf{R}_c$ and translation $\mathbf{t}_c$. The function blends the contributions of all Gaussians overlapping each pixel by accounting for both color $c_i$ and opacity $\alpha_i$. The transmittance $T_i$ makes each Gaussian visible properly, which helps make the rendered image look more realistic. 3DGS utilizes SfM to generate an initial set of points. After that, it goes through steps like densification and pruning to represent the entire scene.

\subsection{Geometric Mapping with Geometric Confidence Score}

The primary goal of GeomGS is to create a highly accurate structural representation based on 3DGS. To achieve this, we first accumulate LiDAR point cloud data using the pose information provided in the dataset. This accumulated data serves as the initial point cloud. Compared to traditional methods like SfM, this approach can produce a denser point cloud. The accumulated point cloud is created by utilizing the transformation matrix \( \mathbf{T}_i \) corresponding to each pose. This transformation matrix converts the LiDAR scan \( \mathbf{P}_i \) into the world coordinate system. The transformed LiDAR scans from all poses are then combined into the final accumulated point cloud \( \mathbf{P} \), as shown in Eq. \ref{accum_lidar}:

\begin{equation}
\label{accum_lidar}
    \mathbf{P} = \bigcup_{i} \left( \mathbf{T}_i \mathbf{P}_i \right)
\end{equation}

To improve the structural accuracy of the map, first, we introduce the Geometric Confidence Score (GCS) \( \gamma \), a new Gaussian parameter that optimizes the identification of geometrically reliable points. For measuring the geometric confidence of a point, we compute using an asymmetric sigmoid function, $\sigma_{\text{asym}}(x)$, as shown in Eq.~\ref{sigmoid}, where $k$ controls the slope, and $d$ determines the dividing point. We set k to 20 and d to 0.9 to distinguish confidence based on the distance between the LiDAR points and Gaussian primitives.
\begin{equation}
\label{sigmoid}
    \sigma_{\text{asym}}(x) = \frac{1}{1 + e^{k (x - d)}} \in [0, 1]
\end{equation}

We optimize the GCS $\gamma$ using the function in Eq.~\ref{sigmoid}. To continuously quantify it during the optimization process, we introduce a new loss term, $\mathcal{L}_{\text{geom}}$, as shown in Eq.~\ref{geom_loss}. This GCS is fully utilized in the process of creating a more precise structure.
\begin{equation}
\label{dist}
    d_i = \min_{\mathbf{p} \in \mathbf{P}} \| \mathbf{g}_i - \mathbf{p} \|^2, \quad \mathbf{g}_i \in \mathbf{G} 
\end{equation}
\begin{equation}
\label{geom_loss}
    \mathcal{L}_{\text{geom}} = \frac{1}{N} \sum_{i=1}^{N} (\gamma_i - \sigma_{\text{asym}} (d_i))^2, \quad \gamma_i \in (0,1)
\end{equation}

Where the term $d_i$ as shown in Eq.~\ref{dist} represents the distance between the $\mathbf{G}$ and its closest accumulated LiDAR point \( \mathbf{p} \) in the $\mathbf{P}$ for all $N$ Gaussian points.

Next, to construct an accurate structural map, we impose probabilistic distance constraints on the Gaussian primitives based on the GCS $\gamma$. Rather than simply minimizing the Euclidean distance, we apply more robust constraints by incorporating GCS, as shown in Eq. \ref{prob_dist}:
\begin{equation}
\label{prob_dist}
    \mathcal{L}_{prob} = \frac{1}{N} \sum_{i=1}^{N} \left( \ln (1-\gamma_i) + \frac{d_i}{(1-\gamma_i)} \right)
\end{equation}
This probabilistic distance optimization method assigns more weight to points that have higher GCS. Focusing more on structurally important points when generating an accurate map, reduces the influence of points that are not structurally crucial (e.g., sky, tall buildings), but are important for image rendering. This approach continuously generates and optimizes an accurate structure without compromising image quality.

In addition, we apply a scale loss \cite{lombardi2021mixture} to prevent the overlapping of Gaussian points, a perceptual loss \cite{johnson2016perceptual} to preserve feature-level details, and an appearance model \cite{lin2024vastgaussian} to enhance robustness against brightness variations. Similar to \cite{lin2024vastgaussian}, $\mathcal{I}_a$ is the image from the appearance model, and $\mathcal{I}_r$ is the rendered image. The final loss design, which includes these components, is shown in Eq.~\ref{rgb_loss} and Eq.~\ref{total_loss}:
\begin{equation}
\label{rgb_loss}
    \mathcal{L}_{rgb} = (1-\lambda_{\text{rgb}})\mathcal{L}_1(\mathcal{I}_a, \mathcal{I}_{gt})+\lambda_{\text{rgb}}\mathcal{L}_\text{D-SSIM}(\mathcal{I}_r, \mathcal{I}_{gt})    
\end{equation}
\begin{equation}
\label{total_loss}
\begin{split}
    \mathcal{L}_{\text{total}} = &\ \mathcal{L}_{\text{rgb}} + \lambda_{\text{geom}}\mathcal{L}_{\text{geom}} + \lambda_{\text{prob}}\mathcal{L}_{\text{prob}} \\
    &\ + \lambda_{\text{scale}}\mathcal{L}_{\text{scale}} + \lambda_{\text{perc}}\mathcal{L}_{\text{perc}}
\end{split}
\end{equation}
The values of the hyper-parameter $\lambda$ used in the loss function are set as follows: $\lambda_{\text{rgb}} = 0.2$, $\lambda_{\text{geom}} = 0.1$, $\lambda_{\text{prob}} = 0.1$, $\lambda_{\text{scale}} = 100.0$, and $\lambda_{\text{perc}} = 0.5$.

\begin{table*}[htbp]
\centering
\begin{threeparttable}
\begin{center}
\caption{Quantitaive Results on KITTI-360 and KITTI}
% \begin{tabular}{lc|ccc|ccc}
% \begin{tabularx}{\textwidth}{l X | X X X | X X X}
\begin{tabularx}{\textwidth}{l >{\centering\arraybackslash}X | >{\centering\arraybackslash}X >{\centering\arraybackslash}X >{\centering\arraybackslash}X | >{\centering\arraybackslash}X >{\centering\arraybackslash}X >{\centering\arraybackslash}X}
\toprule
\multirow{2}{*}{} & \multirow{3}{*}{Initial Points} & \multicolumn{3}{c|}{KITTI-360} & \multicolumn{3}{c}{KITTI} \\
\cmidrule(lr){3-5} \cmidrule(lr){6-8}
 &  & PSNR $\uparrow$ & SSIM $\uparrow$ & LPIPS $\downarrow$ & PSNR $\uparrow$ & SSIM $\uparrow$ & LPIPS $\downarrow$ \\
\midrule

Scaffold-GS \cite{scaffoldgs} & SfM & 18.9783 & 0.7678 & 0.3072 & 18.0842 & 0.6679 & 0.3214 \\ [0.1em]
Scaffold-GS \cite{scaffoldgs} & LiDAR & 20.5806 & 0.7796 & 0.2923 & 18.1341 & 0.6429 & 0.3384 \\ [0.1em]
GaussianSurfels \cite{Dai2024GaussianSurfels} & SfM & 21.0695 & 0.7765 & 0.3569 & 18.4256 & 0.5727 & 0.4791 \\ [0.1em]
GaussianSurfels \cite{Dai2024GaussianSurfels} & LiDAR & 23.0385 & 0.8291 & 0.2844 & 18.3671 & 0.5678 & 0.4796 \\ [0.1em]
2DGS \cite{Huang2DGS2024} & SfM & 23.3374 & 0.8388 & 0.2536 & 21.6865 & 0.7584 & 0.2724 \\ [0.1em]
2DGS \cite{Huang2DGS2024} & LiDAR & \cellcolor{lightyellow}23.6735 & \cellcolor{lightyellow}0.8437 & 0.2459 & 21.5642 & 0.7489 & 0.2845 \\ [0.1em]
3DGS \cite{kerbl3Dgaussians} & SfM & 23.1327 & 0.8231 & 0.2435 & 21.5168 & \cellcolor{lightyellow}0.7611 & \cellcolor{lightred}\textbf{0.2316}  \\ [0.1em]
3DGS \cite{kerbl3Dgaussians} & LiDAR & 23.5725 & 0.8332 & \cellcolor{lightyellow}0.2259 & \cellcolor{lightyellow}21.7329 & \cellcolor{lightorange}0.7622 & 0.2368 \\ [0.1em]
Ours-S & LiDAR & \cellcolor{lightorange}24.2963 & \cellcolor{lightorange}0.8533 & \cellcolor{lightorange}0.2033 & \cellcolor{lightorange}21.8820 & 0.7590 & \cellcolor{lightyellow}0.2367 \\
Ours-P & LiDAR & \cellcolor{lightred}\textbf{24.2981} & \cellcolor{lightred}\textbf{0.8555} & \cellcolor{lightred}\textbf{0.1903} & \cellcolor{lightred}\textbf{21.9829} & \cellcolor{lightred}\textbf{0.7646} & \cellcolor{lightorange}0.2329 \\
\bottomrule
\end{tabularx}
\begin{tablenotes}
    \footnotesize
    \item Evaluate the quality of the rendered image in a conventional test scene.
\end{tablenotes}
\label{table_1_mapping_result}
\end{center}
\end{threeparttable}
\end{table*}

\subsection{Gaussian Splatting-based Unified Localization}

We propose a novel localization method based on a highly accurate map generated from GeomGS. Existing pose optimization techniques, such as iNeRF \cite{yen2020inerf} estimate the pose by minimizing the loss between the rendered output from the current pose and the ground truth image over several iterations. While these methods have demonstrated effective pose estimation, they do not inherently operate within a coordinate system that reflects actual space or accounts for actual scale.

In contrast, our approach leverages the advantages of GeomGS, which allows the use of renderable properties, an accurate Gaussian map, and its confidence scores. The first key idea of our method is to apply Iterative Closest Point (ICP) \cite{original_icp} between the Gaussian points and the query LiDAR scan. Afterward, we iteratively refine the pose by comparing the rendered image at the pose \(\mathbf{R}, \mathbf{t}\), obtained from the ICP results, with the ground truth image. Similar to the approach in iNeRF \cite{yen2020inerf}, as shown in Eq.~\ref{imageref1} and Eq.~\ref{imageref2}. This refined pose is applied to the LiDAR scan \(\mathbf{Q}_p\) again. We repeat these two processes iteratively for a given number of iterations to refine the pose and align the points.

One of the most significant aspects of our approach is the use of the Geometric Confidence Score (GCS), developed within GeomGS, to perform Weighted ICP. GCS assigns reliability scores to each point, and these scores are used as weights for each point pair in the ICP process. This results in more accurate and robust pose estimation.
The core of Weighted ICP is the use of a weight matrix in Eq. \ref{eq:weightform}, which influences the transformation calculation between the source and target point clouds. The relationship between the weighted source points \(\mathbf{S}\) and target points \(\mathbf{T}\) is represented by the matrix \(H\) in Eq. \ref{eq:icp_weighted}, where \(\bar{\mathbf{s}}\) and \(\bar{\mathbf{t}}\) are the centroids (mean points) of the source and target point clouds, respectively. Using Singular Value Decomposition (SVD), the rotation matrix \(\mathbf{R}_{\text{icp}}\) and translation vector \(\mathbf{t}_{\text{icp}}\) are computed, as shown in Eq.~\ref{eq:icp_rotation}:
\begin{equation}
\label{imageref1}
\Delta \mathbf{R}, \Delta \mathbf{t} = \underset{\Delta \mathbf{R}, \Delta \mathbf{t}}{\operatorname{arg\,min}} \left( \mathcal{L} \left(\hat{C}(\mathbf{G} \mid \mathbf{R}, \mathbf{t}) ,\mathcal{I}_{gt} \right) \right)
\end{equation}
\begin{equation}
\label{imageref2}
\mathbf{Q}_p' = \Delta \mathbf{R}  \mathbf{Q}_p + \Delta \mathbf{t}
\end{equation}
\begin{equation}
\label{eq:weightform}
\mathbf{W} = \text{diag}(\gamma)
\end{equation}
\begin{equation} \label{eq:icp_weighted}
\mathbf{H} = (\mathbf{S} - \bar{\mathbf{s}})^\top \mathbf{W} (\mathbf{T} - \bar{\mathbf{t}}), \quad \text{SVD}(\mathbf{H}) = \mathbf{U} \mathbf{S} \mathbf{V}^\top
\end{equation}
\begin{equation} \label{eq:icp_rotation}
 \mathbf{R}_{icp} = \mathbf{V} \mathbf{U}^\top, \quad \mathbf{t}_{icp} = \bar{\mathbf{t}} - \mathbf{R}_{icp} \bar{\mathbf{s}}
\end{equation}

Weighted ICP and image refinement are combined in a way that they support each other as shown in Fig.~\ref{figure_system}, helping to overcome each method's weaknesses. If ICP fails to align the points correctly, image refinement can fix the pose or prevent the process from failing with pixel-level comparison. Likewise, if image refinement struggles with correcting large errors, ICP can help correct them based on accurately structured Gaussian points. This approach takes full advantage of both image rendering and LiDAR-based localization, ensuring reliable localization even in challenging environments.

\section{EXPERIMENTS}

\subsection{Experimental Setup}
We designed our experiments to demonstrate the effectiveness of our system by evaluating the state-of-the-art (i) image rendering quality via GeomGS, (ii) the accuracy of geometric representations under the proposed constraints, which have not been previously evaluated, and (iii) the feasibility and accuracy of our proposed localization method in GeomGS. We selected 100 consecutive images from KITTI \cite{kitti} and KITTI-360 \cite{KITTI360} datasets, representing scenes that cover approximately 100 meters in actual space. Unlike existing methods that rely on pose information obtained from SfM, we used pose data provided by KITTI and KITTI-360. For a baseline comparison, we selected 3DGS, which has achieved SOTA in novel view synthesis.

\subsection{Image Quality Validation}
We evaluated our method using the same approach as 3DGS, with standard metrics such as PSNR, SSIM \cite{ssim}, and LPIPS \cite{lpips}, testing on scenes sampled every 8 frames. However, unlike 3DGS, we set the initial position learning rate to 1.6e-5, instead of 1.6e-4. Additionally, we compared our approach with recent methods like 2DGS \cite{Huang2DGS2024}, Gaussian Surfels \cite{Dai2024GaussianSurfels}, and Scaffold-GS \cite{scaffoldgs}. For a fair comparison, we used ground truth (GT) poses to evaluate these methods, testing them on both the initial SfM points and initial LiDAR points. For comparison with SfM points, we used COLMAP's \cite{schoenberger2016sfm} triangulation to scale the SfM points to the actual size. This process required known camera parameters, which we extracted from the datasets.

Table \ref{table_1_mapping_result} shows that using LiDAR points as initial points improves performance in most methods. While image quality often degrades in complex scenes, initializing with dense LiDAR points helps reduce this issue in most methods. However, the most significant performance improvement is achieved by applying our method, which includes the constraint between LiDAR points and Gaussian points.

\begin{figure*}[tb]
      \centering
      \includegraphics[width=\textwidth]{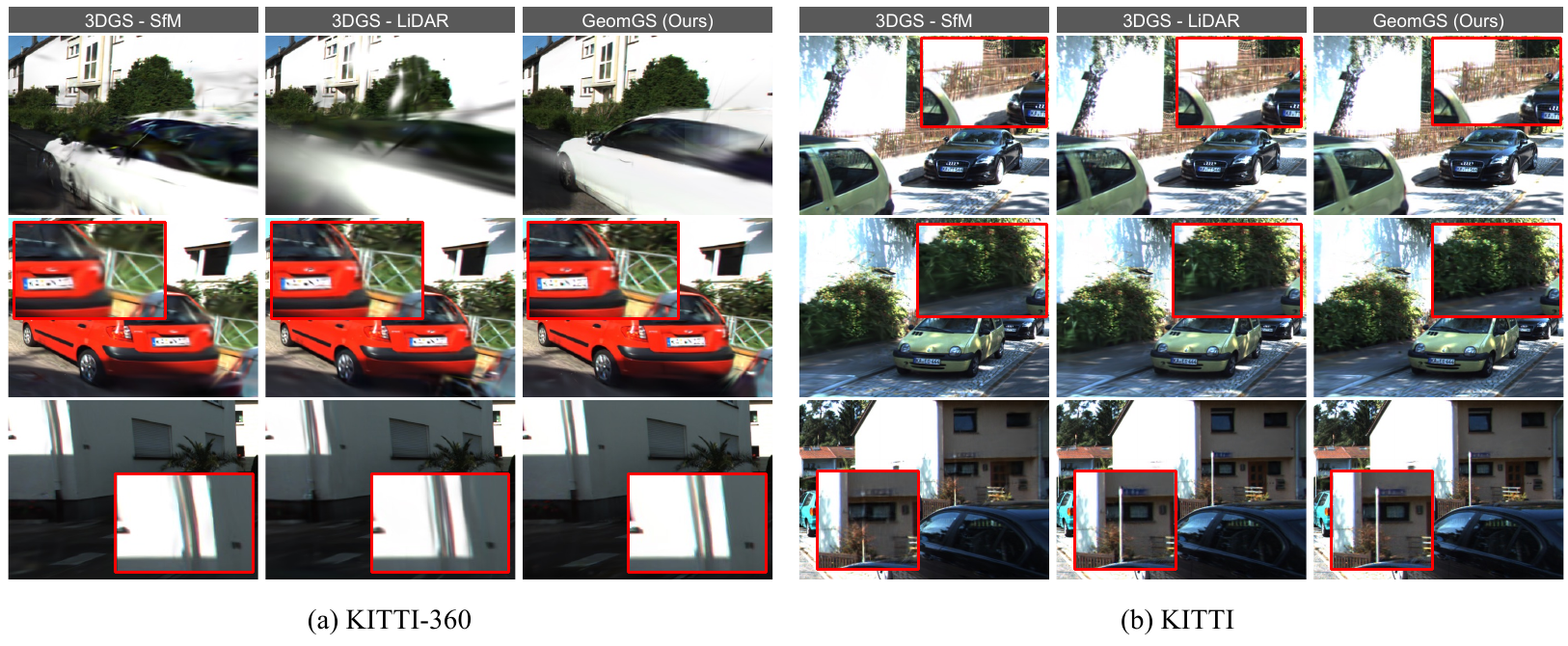}
      \caption{\textbf{Qualitative comparison of GeomGS and 3DGS in (a) KITTI-360 \& (b) KITTI datasets.} Patches represent visually distinct regions, highlighting fine details and geometric variations. Our method performs better in various scenarios by incorporating finer details, improving geometric representation, and enhancing overall image quality. The notation adjacent to 3DGS denotes which specific initial point was utilized in the process.}
      \label{image_quality_figure}
   \end{figure*}

Ours-S applies a simple Euclidean distance loss to our method, whereas Ours-P employs a probabilistic distance loss with GCS. The probabilistic distance loss can also achieve structural characteristics without degrading image performance. Fig. \ref{figure_one} and Fig. \ref{image_quality_figure} illustrate that our method captures fine details and structural accuracy more effectively across various scenes. It also successfully addresses shape distortion in novel views.

\begin{table}[h!]
\centering
\caption{Geometric Performace on KITTI-360}
\begin{tabular}{lc|cccc}
\toprule
& \makecell{Initial \\ Points} & \makecell{F-Score $\uparrow$ \\ @0.1} & \makecell{F-Score $\uparrow$ \\ @0.2} & \makecell{F-Score $\uparrow$ \\ @1.0} & CD $\downarrow$ \\ 
\midrule
3DGS & SfM & 0.4220 & 0.4912 & 0.6393 & 261.1804 \\ [0.2em]
3DGS & LiDAR & 0.5639 & 0.6456 & 0.8132 & 6.6338 \\ [0.2em]
2DGS & SfM & 0.4243 & 0.4953  & 0.6589 & 183.5640  \\ [0.2em]
2DGS & LiDAR & 0.4783 & 0.5550  & 0.7153 & 91.7571  \\ [0.2em]
Ours-S & LiDAR & 0.8230 & 0.8822 &  0.9534 & 2.8688 \\
Ours-P & LiDAR & \textbf{0.8948} & \textbf{0.9267} &  \textbf{0.9629} & \textbf{2.6709} \\ 
\bottomrule
\end{tabular}

\label{table:results_fscore}
\end{table}

\begin{table*}[htbp]
    \caption{Comparison of localization methods based on rotation and translation errors}
    \centering
    \begin{threeparttable}
    \begin{tabularx}{\textwidth}{c *{5}{>{\centering\arraybackslash}X}}
        \toprule
        \textbf{Initial Error} & \textbf{ICP (SfM)} & \textbf{ICP} & \textbf{iNeRF \cite{yen2020inerf}} & \textbf{WICP} & \textbf{Ours} 
        \\ \midrule
        20.0 / 2.0 & 2.8548 / 1.7532 & \textbf{0.7388} / 0.9853 & 15.4341 / 1.2635 & 4.0217 / 1.0654 & 0.8635 / \textbf{0.5396} \\ [0.2em]
        20.0 / 3.0 & 6.7209 / 6.4538 & 6.2766 / 5.2994 & 22.5281 / 3.2002 & 3.9178 / \textbf{1.1412} & \textbf{3.8972} / 1.5464 \\ [0.2em]
        30.0 / 2.0 & 11.9116 / 24.4708 & 11.1796 / 7.4975 & 37.1290 / 2.3754 & \textbf{6.9021} / \textbf{0.9786} & 7.0102 / 1.1784 \\ [0.2em]
        30.0 / 3.0 & 15.0402 / 5.7775 & 16.0031 / 4.6866 & 34.6142 / 3.2958 & 9.8547 / \textbf{1.6552} & \textbf{3.2957} / 2.3108 \\ [0.2em]
        40.0 / 4.0 & 22.2579 / 8.9544 & 20.4090 / 9.3045 & 45.4746 / 3.8619 & 15.2049 / 3.1879 & \textbf{14.6590} / \textbf{2.7048} \\ [0.2em]
        \bottomrule
    \end{tabularx}
    \begin{tablenotes}
        \footnotesize
        \item The values in the table are presented as ``rotation error [$^\circ$] / translation error [m]"
    \end{tablenotes}
    \end{threeparttable}
    \label{tab:localization_table}
\end{table*}

\subsection{Geometric Quality Validation}
We impose initial LiDAR Points and apply constraints between Gaussian points and LiDAR points to enhance the accuracy of reconstruction. To evaluate the structural improvement over existing methods, we measure the similarity between the two point clouds, following the approach in Points2NeRF~\cite{zimny2024points2nerf}. Specifically, we calculate Chamfer Distance (CD), as shown in Eq. \ref{chamfer}, and the F-Score, as shown in Eq.~\ref{fscore}. We then compare these metrics between the generated Gaussians $\mathbf{G}$ and the accumulated LiDAR Points~$\mathbf{P}$.

It is important to note that the reconstructed Gaussian points cannot perfectly match the initial points, especially in areas where LiDAR data lacks coverage, such as the sky or tall buildings. Nevertheless, we observe significant improvements in both CD and F-Score, as shown in Table~\ref{table:results_fscore}. Simply replacing the initial points with Gaussians leads to better results, but our approach using constraints between the points achieves superior reconstruction accuracy. Similarly, our approach using probabilistic distance loss (Ours-P) with GCS achieves an outstanding structure.
\begin{equation}
\label{chamfer}
\begin{split}
    CD(\mathbf{P_1}, \mathbf{P_2}) = & \frac{1}{|\mathbf{P_1}|}\sum_{\mathbf{p} \in \mathbf{P_1}} \min_{\mathbf{q} \in \mathbf{P_2}} \| \mathbf{p} - \mathbf{q} \|^2 \\
    & + \frac{1}{|\mathbf{P_2}|}\sum_{\mathbf{q} \in \mathbf{P_2}} \min_{\mathbf{p} \in \mathbf{P_1}} \| \mathbf{q} - \mathbf{p} \|^2
\end{split}
\end{equation}
\begin{equation}
\label{precision}
\begin{aligned}
        & \text{precision}_1 = \frac{1}{|\mathbf{P_1}|} \sum_{\mathbf{p} \in \mathbf{P_1}} \mathds{I}\left( \min_{\mathbf{q} \in \mathbf{P_2}} \|\mathbf{p} - \mathbf{q} \|^2 < \tau \right), \\
        & \text{precision}_2 = \frac{1}{|\mathbf{P_2}|} \sum_{\mathbf{q} \in \mathbf{P_2}} \mathds{I}\left( \min_{\mathbf{p} \in \mathbf{P_1}} \|\mathbf{q} - \mathbf{p} \|^2 < \tau \right)
\end{aligned}
\end{equation}
\begin{equation}
\label{fscore}
    \text{F-score} = \frac{2 \times \text{precision}_1 \times \text{precision}_2}{\text{precision}_1 + \text{precision}_2}
\end{equation}

\subsection{3D Localization Performance}
We evaluate the 3D localization performance on our structurally accurate map. For the evaluation, we selected test cases from every 10th sequence out of 100 sequences and calculated the average localization performance. And we intentionally applied large initial errors to poses. Our method performs localization by iteratively leveraging the properties of the structurally accurate map and its image rendering capabilities. In particular, we utilize the Geometric Confidence Score (GCS) as the weight in the Weighted ICP (WICP) algorithm to enhance its robustness and accuracy. We compare our approach with existing methods such as ICP \cite{original_icp}, iNeRF \cite{yen2020inerf}, and our WICP.
\begin{figure}[tb]
  \centering
  \includegraphics[width=\columnwidth]{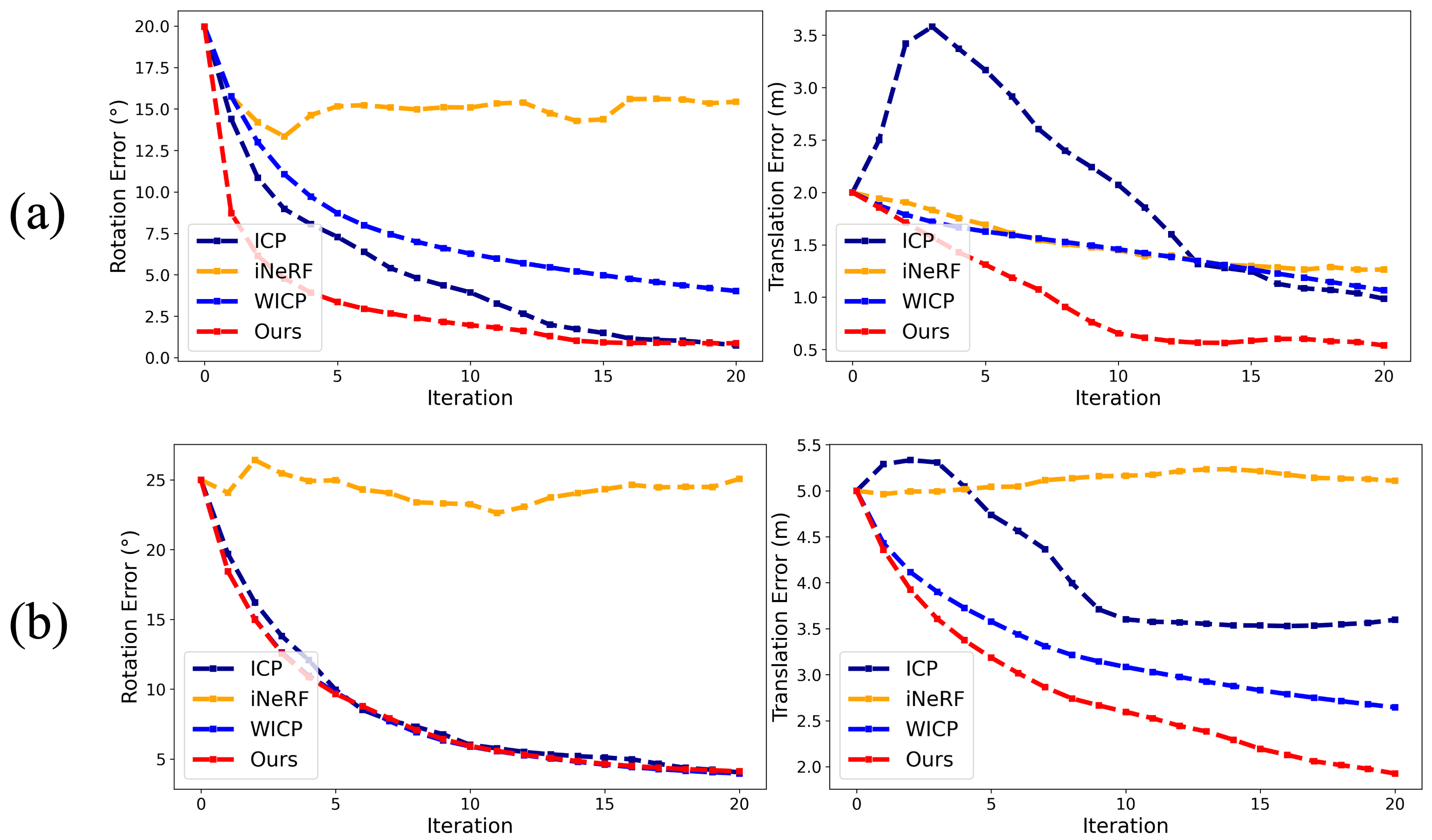}
  \caption{Comparison of ICP \cite{original_icp}, WICP, Image Refinement (iNeRF \cite{yen2020inerf}), and Ours per Iteration. (a) Initial error : 20.0$^\circ$ / 2.0m, (b) Initial error : 25.0$^\circ$ / 5.0m}
  \label{figure_loc_graph}
\end{figure}
Our approach uses 20 iterations in total. In each iteration of our method, WICP is applied once, followed by image refinement over 20 steps. To ensure a fair comparison, the same number of iterations is applied to the other methods. WICP generally performs well on our map, and overall, our proposed method shows strong performance. We evaluate both rotation and translation errors, and as shown in Table \ref{tab:localization_table}, our method typically produces better results. Fig. \ref{figure_loc_graph} illustrates the error reduction across iterations, where our method generally demonstrates superior performances.

\section{CONCLUSIONS}
We propose GeomGS, a method for representing environments with accurate structures and enabling precise localization. We introduce a Geometric Confidence Score (GCS) to identify the geometric reliability of each point. Using a probabilistic distance optimization approach based on GCS, we generate more precise structures without degrading image quality. For localization, we present a novel approach that leverages GCS, LiDAR-based localization, and 3DGS rendering within our accurate map. We evaluate both qualitatively and quantitatively how our maps preserve structural accuracy without compromising image quality and analyze localization performance using these maps.

\newpage

\addtolength{\textheight}{-12cm}  

\bibliographystyle{IEEEtran}
\bibliography{ref}

\end{document}